# Novel Concept-Oriented Synthetic Data approach for Training Generative AI-Driven Crystal Grain Analysis Using Diffusion Model

A. S. Saleh[1,2], K. Croes[2], H. Ceric[3], I. De Wolf[1,2], H. Zahedmanesh[2]

[1] Dept. Materials Engineering, Fac. Engineering Sciences, KU Leuven, B-3001, Leuven, Belgium

[2] imec, Kapeldreef 75, B-3001, Leuven, Belgium

[3] Institute for Microelectronics, TU Wien, Gußhausstraße 27–29/E360, 1040 Wien, Austria

**Abstract**

The traditional techniques for extracting polycrystalline grain structures from microscopy images, such as transmission electron microscopy (TEM) and scanning electron microscopy (SEM), are labour-intensive, subjective, and time-consuming, limiting their scalability for high-throughput analysis. In this study, we present an automated methodology integrating edge detection with generative diffusion models to effectively identify grains, eliminate noise, and connect broken segments in alignment with predicted grain boundaries. Due to the limited availability of adequate images preventing the training of deep machine learning models, a new seven-stage methodology is employed to generate synthetic TEM images for training. This concept-oriented synthetic data approach can be extended to any field of interest where the scarcity of data is a challenge. The presented model was applied to various metals with average grain sizes down to the nanoscale, producing grain morphologies from low-resolution TEM images that are comparable to those obtained from advanced and demanding experimental techniques with an average accuracy of 97.23%.

The microstructure of polycrystalline materials dictates their mechanical-, thermal-, transport-, and electrical properties [1], making it crucial to correctly determine, understand, and predict microstructures morphology. By examining features such as grains sizes, grains orientation distribution, and defects, researchers can gain insights into material performance and reliability. This is essential for optimizing material design and manufacturing processes, ensuring that materials meet the stringent requirements for applications ranging from aerospace to electronics. Especially for nano-interconnects in the field of nanoelectronics [2], where the trend towards miniaturization necessitates a deeper understanding of how microstructural features influence material reliability affected by various failure mechanisms [3]. Additionally, properties such as electrical resistivity and thermal conductivity play a pivotal role in determining performance, with the goal of enabling rapid signal transmission and reduced power consumption in nano-interconnects [4].

Historically, the extraction of grain boundaries from images produced by any microscopy technique was conducted manually, limiting the scope, objectivity, fidelity and efficiency of microstructural analysis [5]. Early automation efforts in the late 20th century utilized basic image processing techniques like thresholding and edge detection [6] [7]. Advances in computational power during the 1990s and 2000s enabled more sophisticated methods, including using gradient-based and morphological filters such as Sobel [8] and Canny [9] edge detectors, although their direct application to transmission electron microscopy (TEM) images often fell short due to noise [10] [11]. Noise can be defined as any variation in the image obscuring true information. These variations can be either fully stochastic or following a certain pattern depending on their source. Numerous sources of noise can affect microscopy images such as shading, lack of focus, instrumental and environmental factors [12]. Innovative image analysis approaches, such as the techniques used by Campbell et al. (2018) involving filtering and watershed transforms, improved the automatic delineation of grain boundaries in scanning electron microscopy (SEM) and optical microscopy (OM) images [13] but without achieving the required high levels of accuracy. High-Resolution Transmission Electron Microscopy (HRTEM), offers unparalleled atomic-level imaging, making it ideal for mapping crystals orientation, and defining grain boundaries using Precession Electron Diffraction (PED). However, HRTEM comes with significant drawbacks: it requires complex and labour-intensive sample preparation, and the instrumentation itself is expensive, often necessitating specialized facilities. Moreover, due to the high cost of maintaining and operating HRTEM equipment, its accessibility can be limited, making it a less feasible option for routine or large-scale studies.

The integration of machine learning (ML) in the 2010s and the 2020s marked a significant shift towards advanced automation in image segmentation, with convolutional neural networks (CNNs) demonstrating superior performance in image analysis [14] [15]. Recent advances in computational power have enabled the development of increasingly complex machine learning models with deeper architectures and a higher number of learnable parameters. Especially huge leaps of advancement being witnessed in the development of generative artificial intelligence (AI) models. Specifically, diffusion models (DMs), which learn the probability distribution of a given training dataset by simulating a diffusion process [16] [17], where successive loops of noise removal can achieve an output of high quality, making DMs an attractive option to many different fields of applications [18] [19]. This growth, however, has exacerbated the challenge of data scarcity, as more data are required to effectively train these sophisticated models [20]. This challenge can be a limiting factor for the performance of such ML models in areas of research where sufficient clear labelled data for training are not available, like in the case of automated microstructure analysis from microscopy images.

This study addresses the data scarcity issue by introducing a novel recipe for generating concept-oriented synthetic data, given in details in the methods section. This approach not only supplements the available data but also provides enhanced control over model training. Throughout this study the word "*Concepts*" is used to indicate the rules and ideas required to be known to the ML model in order to be able to achieve the task it is designed to accomplish after its training. For example, a ML model designed to predict the electric field lines in the presence of conductors would need to understand that field lines come out from positively charged bodies and enter negatively charged bodies [21], there is no electric field inside a charged conductor [21], the electric field is always perpendicular to the surface of a conductor in electrostatic equilibrium [21], and higher curvature on the surface of the conductor causes higher density in electric field lines [21] [22]. Another example for concepts in the field of classical mechanics would be the three Newton's laws of motion [23], which would then be essential for a ML model designed to handle such mechanical problems. By incorporating the appropriate concepts and adjusting noise severities, the synthetic data are tailored to optimize model behaviour, offering advantages over real-world data that often contains excessive, uncontrolled noise. The use of synthetic data is similar to educational strategies where simplified, concept-driven problems generated by the instructor, help learners grasp complex ideas. Similarly, this approach demonstrates that synthetic data can effectively guide machine learning models, facilitating improved training outcomes and model performance.

In this study, we aim to surpass current methods for automated microstructure extraction through the integration of advanced image processing with modern diffusion models and encoder-decoder architectures trained using concept-oriented synthetic data. Results show a high-accuracy microstructural analysis comparable to advanced and more demanding microscopy techniques for crystal structure mapping.

## Creating synthetic dataset for training

Before setting the model components and architecture, the dataset should be first generated, which will then be used to train the model. A summary of the process of generating dataset images is illustrated in Fig. 1a. As shown in the figure, the synthesis of binary images serving as target for training, was achieved using our custom Monte Carlo generator module. This module takes in a predefined grain size probability distribution, then iteratively draws circles of various diameters from this probability distribution until the total area of the circles drawn matches the required image size. To achieve optimal packing, a dynamic packing algorithm is applied, incorporating random perturbation forces to ensure minimal overlap, and reducing interstitial spaces between the circles. Unlike conventional methods, such as random placement techniques which add circles based on remaining space [24], our approach maintains the specified size distribution of all circles from the outset. Following circle generation, we employed a weighted Voronoi tessellation, also known as power diagram [25], to form the actual grains. The centres of the circles serve as the seeds for the Voronoi diagram, while their radii provide the weights. Then, the grain boundaries were delineated with a specified thickness. Pixels within this boundary thickness were assigned a value of 1, while all other pixels were set to 0, thus producing the desired binary image (see Fig. 1a). Finally, to create the synthetic TEM images to be used as an input during model training, each grain in the binary images was assigned a random gray scale intensity, after which appropriate types and levels of noise were introduced. The details of how concept-oriented synthetic data approach was used to create this data generator along with the details of the training process is given in the methods section of this study.

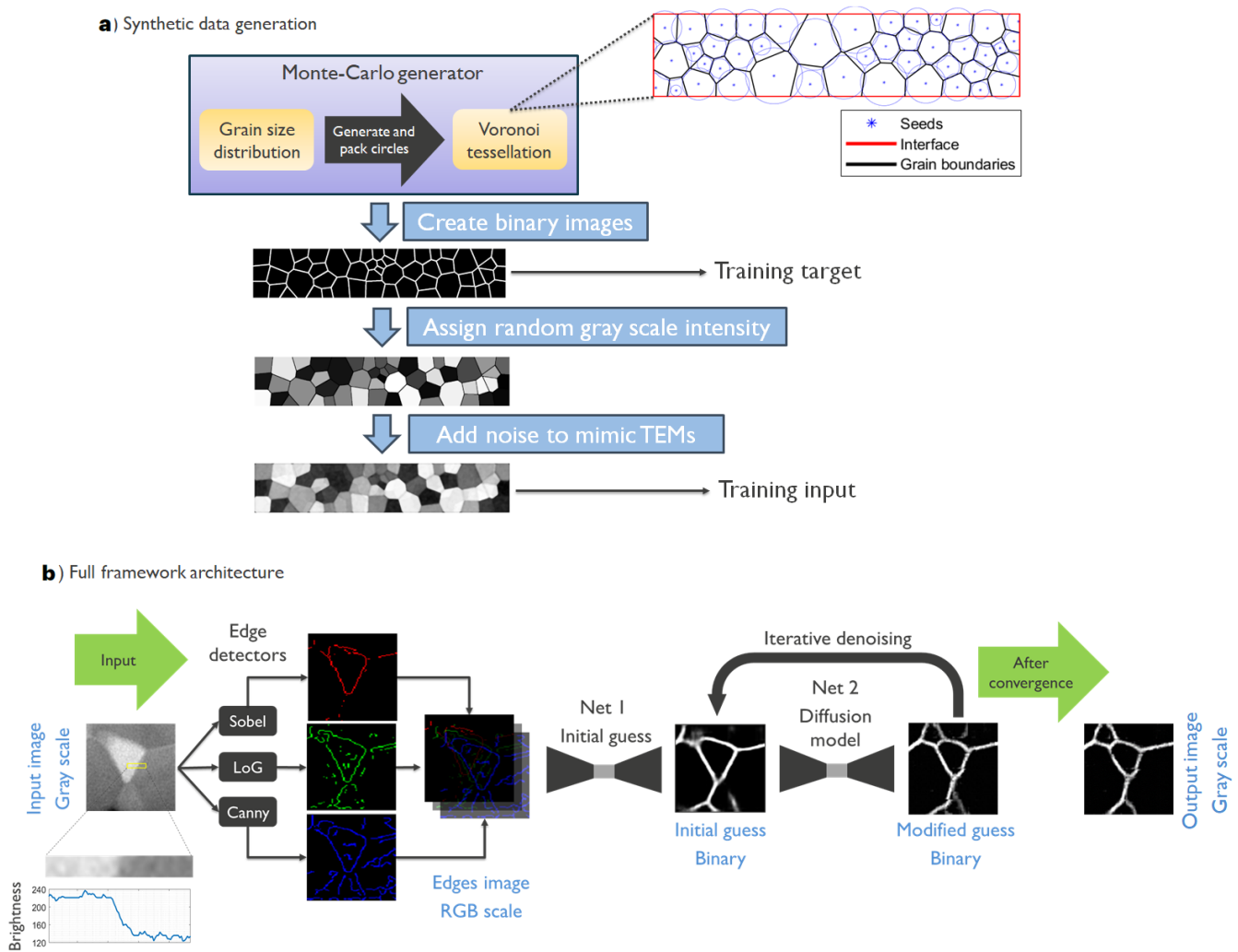

Figure 1: **Dataset generation and model components summary**. (a) Monte-Carlo generator employs Voronoi tessellation, assigning random grey scale color to each grain, and adding overlap effect and noise to generate synthetic Target/Input data for training following the concept-oriented procedure. (b) The full architecture of the presented model showing the three main stages and the In/Out data type for each. An example from a TEM image is given to show how grain boundaries are usually visible as borders between regions with different brightness in microscopy images due to the different scattering behavior each crystal orientation has on incoming beams.

**Model components**

Our automated microstructure analysis framework consists of three key components: (i) Edge detectors, (ii) Initial guess network, and (iii) Generative modifier using the diffusion model (See Fig. 1b). Edges are defined to be the set of curves within an image at which variations in intensity, colour, or texture occur [26]. These curves are most likely to match objects boundaries, however they can also be detected due to noise and other artefacts in the image. An edge detector with high sensitivity would capture all edges including the ones coming from noise and not matching an object boundary [27] while a low sensitivity edge detector would only detect the edges at strong variations in intensity and ignore the ones with weak variations to filter noisy edges, however this comes at the risk of ignoring edges at objects boundaries if the boundary separates two regions of close intensity or colour [27]. In the edge detection stage of the presented model, the input grayscale image is processed using three different edge detection techniques to capture a comprehensive range of edges features. The first technique utilizes a gradient-magnitude edge detector, specifically the Sobel operator [8], which highlights intensity gradients. The second technique employs a zero-crossing detector based on the Laplacian of Gaussian (LoG) [28], providing sensitivity to variations in curvature. The third technique is the Canny edge detector [9], known for its robustness in identifying fine edges. The outputs from these edge detectors are compiled into a single three-channel RGB image (i.e., n×m×3 matrix), where n and m are the image width and length respectively and the third dimension is the red, green and blue channels of the image with each channel representing one of the edge detection maps. By integrating these methods, with their varying levels of sensitivity, a detailed representation of edges is generated, aiming to enhance the subsequent stages' ability to accurately distinguish genuine grain boundaries from noise. The second stage involves the application of an encoder-decoder neural network, represented by Net1 in Fig. 1b, where the RGB image from the edge detection stage is the input and a binary classification image is the output. In this binary image, pixels with a value of 1 correspond to grain boundaries, while pixels with a value of 0 correspond to the grain bulk. This output serves as an initial guess for the final stage of the process. The final stage employs another encoder-decoder neural network that operates with binary images of the same size as input and output. This network functions as a diffusion model, represented by Net2 in Fig. 1b, where each iteration refines the image by using the previous iteration's output as the next iteration's input. The objective of these iterations is to progressively reduce noise and connect any fragmented grain boundaries, thereby enhancing the quality of the microstructure extraction. The iterative process continues until convergence, where no significant changes are observed between successive iterations. A convergence criterion is established through sensitivity analysis, whereby the model ceases iterations when the sum of the absolute differences between pixel values at the start and end of an iteration is smaller than 0.003 times the product of the total number of pixels and the maximum pixel value (i.e., 255).

## Results

A step-by-step demonstration of how the proposed system extracts microstructures from a TEM image is provided from top to bottom in Fig. 2a. The three stages—edge extraction, initial microstructure guess, and final denoising with the diffusion model—are thoroughly illustrated.

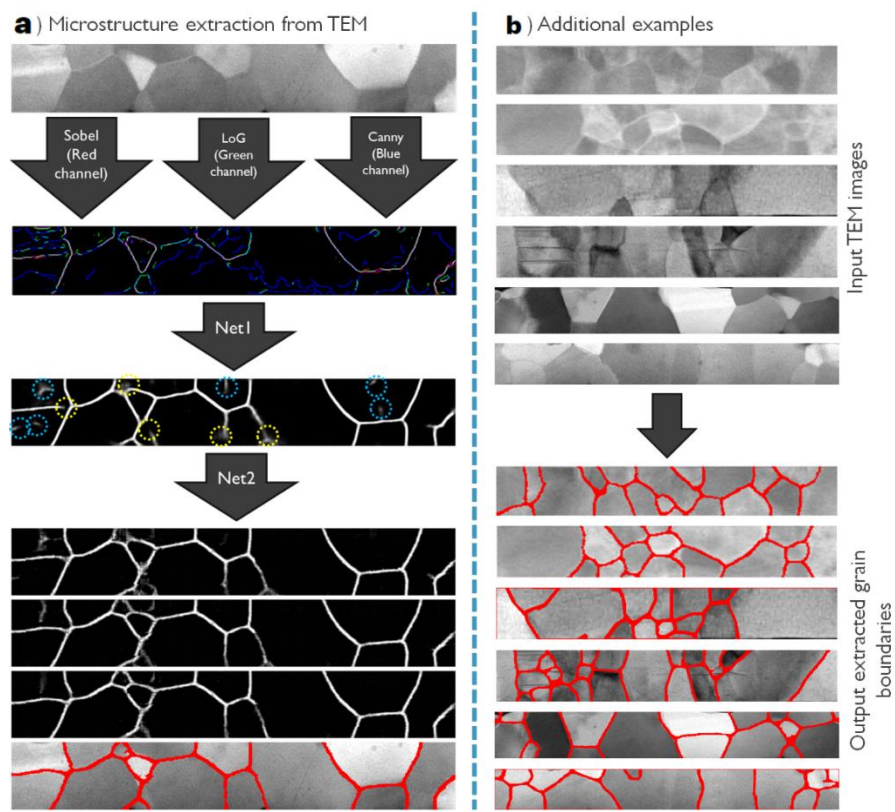

*Figure 2: **Microstructure extraction from TEM images.** a) A step-by-step demonstration of the microstructure extraction process from a TEM image of a Ru line. Residual noise is annotated by blue circles and broken grain boundaries are annotated by yellow circles. The residual noise fragments don't lie on a true grain boundary whereas broken grain boundaries do. The final stack of images shows four equidistant images from a total of 7 iterations by the trained diffusion model, where the final image with resolved grain boundaries in red is overlapped on top of the input TEM for comparison. b) Additional examples for various metal interconnects with resolved grain boundaries in red. First four examples from the top are for Cu lines while the last two are for Ru lines.*

The image analysis starts with the aforementioned edge detectors: Sobel, Laplacian of Gaussian (LoG), and Canny, each with its own sensitivity level giving the RGB image to Net1. The primary function of Net1 is to preserve edges where agreement exists among all three or at least two of the edge detector methods, signifying robust and clear grain boundaries. For edges detected exclusively by the high-sensitivity filter (i.e., the blue edges from the Canny method), the network evaluates their context to decide whether to retain or discard them. Specifically, edges positioned between clear grain boundaries and aligned to form a closed convex polygon are likely true grain boundaries with weak brightness gradients and should be retained. Conversely, isolated edges that do not connect to strong ones are likely artifacts of noise and should be excluded. This initial guess is then iteratively processed by the diffusion model (Net2) to refine grain boundary detection by eliminating any residual noise not filtered by earlier stage (annotated by blue circles in Fig. 2a) and connecting fragmented boundaries (annotated by yellow circles in Fig. 2a). The final comparison between the extracted microstructure and the TEM image, demonstrates the framework's capability to produce a well-connected and clear microstructure that closely matches the TEM observations. Additional examples for applying the whole framework on different metal interconnects are shown in Fig. 2b. demonstrating the model's versatility and applicability across a wide range of imaging conditions and materials.

## Model benchmarking

To assess the accuracy of the presented model, its output is compared to those obtained using advanced experimental methods for characterizing grain structures. Specifically, results from a state-of-the-art study on the microstructure of Molybdenum (Mo) are used [29]. This study provided both TEM images and their corresponding inverse pole figure (IPF) maps constructed from electron backscatter diffraction (EBSD) patterns and validated using X-ray diffraction (XRD). These comprehensive datasets were ideal for evaluating our model's accuracy in extracting microstructures from TEM images and comparing these with the boundaries visible in the IPF maps representing the ground truth. The results for three different case studies are illustrated in Fig. 3, which includes the following: (i) An unmarked input TEM image, (ii) the corresponding unmarked IPF map. (iii) The TEM image annotated with grain boundaries extracted by our model marked in red. (iv) The IPF map annotated to show grain boundaries correctly detected by our model marked in white, and those missed marked in orange.

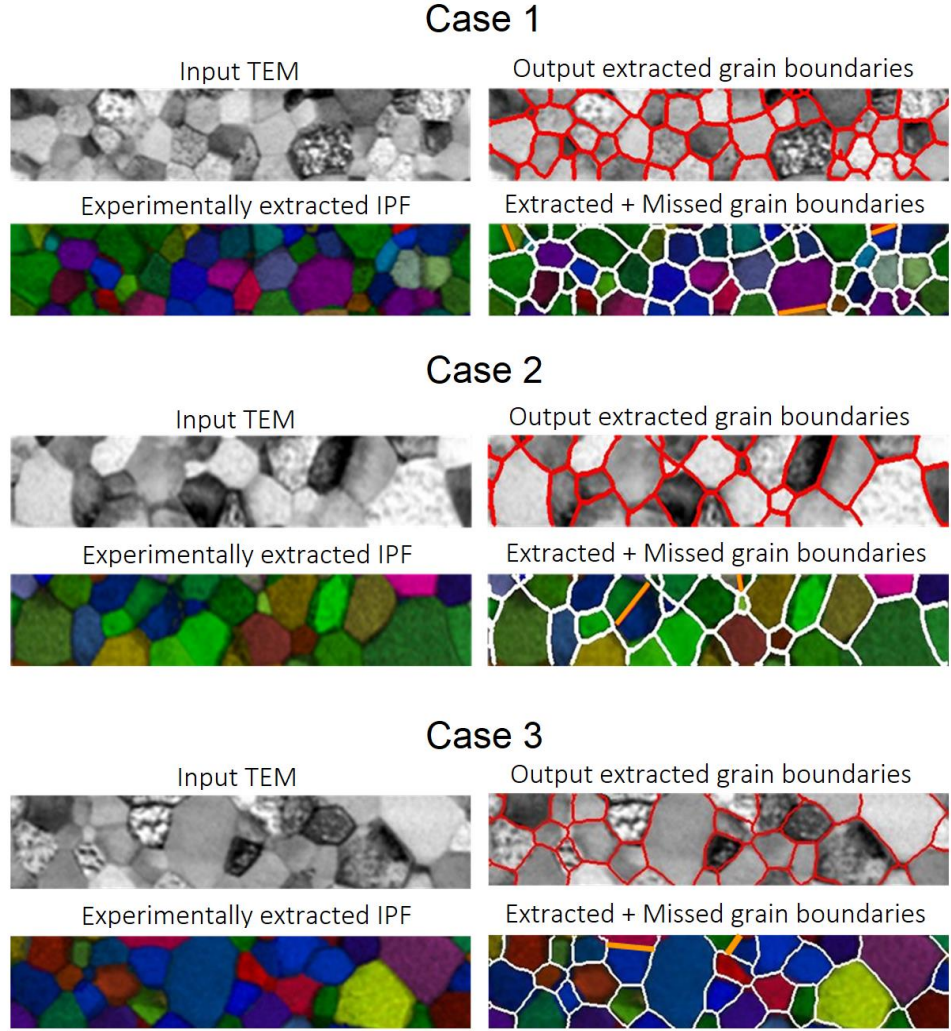

*Figure 3: **Model accuracy evaluation**. A comparison between microstructure extracted from three different TEM examples of Mo (top left image in each case) to their corresponding experimentally extracted IPF maps using EBSD and XRD adapted from ref. [29] (bottom left image in each case). The grain boundaries extracted by the presented model are annotated in red on top of the input image (top right image in each case). The grain boundaries extracted by the presented model are annotated in white on top of the IPF maps along with the missed grain boundaries marked manually in orange (bottom right image in each case).*

The accuracy metric for each example is defined by Eq. (1), as the percentage of the length of correctly extracted grain boundaries to the total length of all grain boundaries, given a uniform grain boundary thickness.

$$Accuracy \% = 100 \times \frac{No.\,of\,pixels\,in\,correctly\,detected\,boundaries}{Total\,No.\,of\,pixels\,in\,all\,grain\,boundaries} \quad (1).$$

This metric provides a quantifiable measure of our model's performance in detecting and delineating microstructure. The accuracy score for each case is shown in table 1. An average accuracy of 97.23% is achieved across the three cases. This constitutes a significant improvement, exceeding the reported accuracy levels of some state-of-the-art software-based approaches for automated microstructure extraction, such as UNet+CHAC, which has a reported accuracy level of 89% [15].

Table 1: *Accuracy metric for each case in Fig. 3*

| Case | Accuracy |
| --- | --- |
| Case 1 | 97.2% |
| Case 2 | 97.1% |
| Case 3 | 97.4% |

## Discussion

In this study, the edge detectors and the following Encoder/Decoder network (Net1) are employed in the beginning of the presented framework to produce an initial guess for the microstructure which is then given to the diffusion model (Net2) (see Fig. 1b) to predict missing data. This is different to the conventional usage of diffusion models as generators of new data from pure noise [17] [30] [31] [32]. To shed light on this difference, the trained diffusion model can be thought of as a tool to minimize an error function that quantifies the deviation of a structure from a conceptually feasible microstructure, such as the microstructures used in the model training. Each iteration through the diffusion model progressively refines the image, bringing it closer to such a conceptually feasible microstructure until convergence is achieved. In this context, pure noise is characterized by a high error level because it bears no resemblance to a microstructure while the output structures post-convergence must reside within a low-error level as visualized by the sketch in Fig. 4. The true microstructure captured by TEM must lie within the low-error level as well. The results in Fig. 4a show that the presented trained model successfully generated a microstructure from pure noise, indicating that it is possible to do effective generative models training using concept-oriented synthetic data. There is a vast array of possible structures that the diffusion model could generate from noise, each requiring numerous iterations. This introduces significant variability in the output with any slight variation in input, however, to extract the correct microstructure in the input image this range of possibilities must be narrowed and directed toward the specific structure in the low-error level that corresponds to the actual microstructure in the input image.

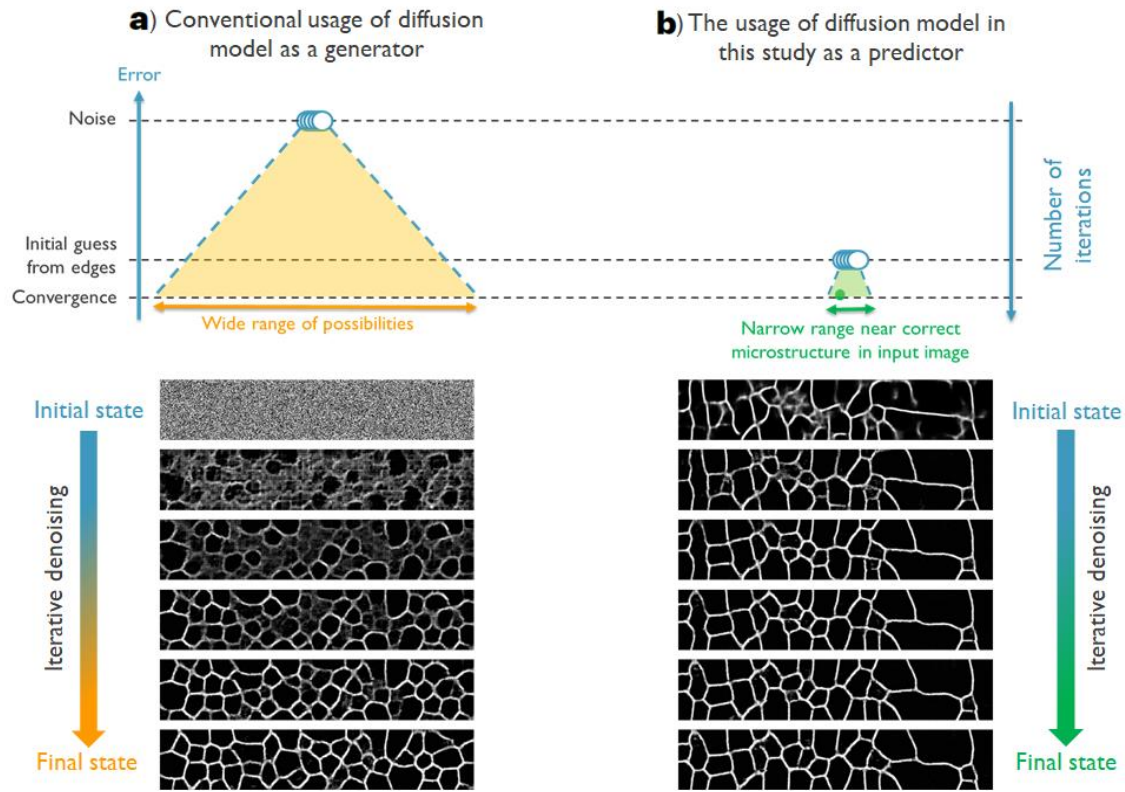

*Figure 4: **impact of changing the initial state of diffusion models**. A sketch to visualize the difference between the conventional usage of a diffusion model as a generator of new data from noise and its usage in this work as a predictor for missing information. a) The case of using pure noise as initial state causing huge variation in generated possibilities for slight variations in the input. The images shown from top to bottom are at the beginning of 6 equidistant loops of iteration out of a total of 46. b) The case of using detected edges from input image as initial state for the diffusion model resulting in significant reduction in the number of possibilities approaching the desired microstructure to extract. The images are at the beginning of 6 equidistant loops of iteration out of a total of 28.*

The initial guess produced by the first two stages in the framework (i.e., Edge detectors and Net1) represents a structure that approximates the required microstructure with some level of distortion, which depends on the input image's quality and the microstructure's clarity. Distortions typically include broken grain boundaries that need reconnecting and isolated edges that should be removed. This initial guess, puts the initial state of the diffusion model closer to the low-error level where the required/actual microstructure exists, thereby reducing the number of iterations required for the diffusion model to converge. Consequently, this minimizes the number of possible structures that the model could generate, ensuring that the final output comes close to the desired microstructure as visualized in Fig.

4b. This approach opens up new possibilities for utilizing diffusion models to predict missing data, starting from an initial approximation.

**Methods**

The methods section in this study is composed of four subsections ordered as follows. The first subsection illustrates the seven-stage methodology of generating concept-oriented synthetic data in a general sense to be applied to many different fields of interest. The second subsection shows its application in generating the synthetic TEM images used for the training of the neural networks shown in this study for microstructure extraction. The pipeline and architectures of these networks are explained in the third subsection with the details of their layers. Finally, in the fourth subsection the details of the training process are listed.

A. Concept-oriented synthetic data

The concept-oriented synthetic data generation methodology presented in this study is structured as follows:

1) **Identification of Data Requirements**: Establish the types of input and target data needed for training (e.g., images, sound, text).
2) **Concepts Listing**: Enumerate all necessary concepts that the machine learning (ML) model must comprehend to accomplish the intended task and devise methods to encode each concept within the data types identified in step 1.
3) **Dimensionality Estimation**: Determine an appropriate data dimensionality that is sufficiently large to encompass all relevant concepts in each example of the generated dataset, yet adequately small to ensure the ML model's complexity remains manageable within the constraints of available computational resources.
4) **Features Probability Distribution**: Identify the primary high-level features that characterize the data and assign a probability distribution to each feature.
5) **Monte Carlo-Based Data Generation**: Utilize a Monte Carlo generator to produce diverse training target examples in accordance with the probability distributions defined in step 4.
6) **Concept Encoding**: Encode all specified concepts into each generated training target example to transform it into a training input example, using the methods devised in step 2.
7) **Noise Introduction**: Add appropriate levels and types of noise to generated input data from step 6.

B. Application to Synthetic TEM Image Generation

Applying the aforementioned methodology to generating synthetic TEM images, produces both input and target data used in training process. The input data are grey scale images, and the target data are binary images, wherein, a pixel value of 1 denotes a grain boundary, and a value of 0 denotes the grain bulk (step 1). Following step 2, the key microstructural concepts to be imparted to the ML model are:

(i) Grains are represented as convex polygons, reflecting the energy minimization during crystallization processes [33] [34].

(ii) Grain boundaries typically occur between regions with different brightness levels, with a lower likelihood of appearing in areas with similar brightness due to the symmetry of crystal structures where differently oriented grains may produce similar scattering behaviour to incoming beams, hence, similar brightness levels in TEM images [35].

(iii) Faint grain boundaries may arise in TEM images due to the overlap of grains at different depths within the sample, these should be disregarded during microstructure extraction [35].

For our application, a suitable data dimensionality for synthetic TEM images of nano-interconnects is chosen to be a single-channel grayscale image with dimensions of 96x496 pixels (step 3), since this size ensures that all concepts are encoded evenly in every example. Although numerous features could describe the microstructure, for simplicity, we focus on grain sizes as the primary high-level feature. Extensive studies have shown that grain size distributions typically follow a lognormal distribution (step 4) [2] [36]. A total of 3000 binary images with varied average grain size were created using our custom Monte-Carlo generator, described earlier in this study (step 5). These binary images were used as target for the training process. Steps 6 and 7 are next used to produce from each target image a corresponding input image that mimics a real TEM of a microstructure. Encoding concept (i) was achieved because a Voronoi tessellation approach was used, which produces convex polygons representing grains [37]. To encode concept (ii), each pixel group belonging to a specific grain was assigned a random grayscale value (See Fig. 1a) within the range of unsigned 8-bit integers (0 to 255). This method ensured that brightness gradients around grain boundaries were accurately represented, including the very small gradients occurring due to the cases where symmetric crystals share a grain boundary. This symmetry makes crystals of different orientations scatter electrons similarly and appear with the same level of brightness under TEM. To encode concept (iii), each image was duplicated and flipped both horizontally and vertically. This flipped copy was then multiplied by a weakening factor of 0.1 and added to the original unflipped image to simulate the overlapping effect of grains at different depths (See Fig. 1a). Since the training target images only contain the original grain boundaries, the model learns to disregard these faint, overlapped boundaries during training. Finally, following step 7, three types of noise were introduced to the images, "Salt & Pepper" noise with a density of 5% of the total pixels, zero-mean "Gaussian" white noise with a variance of 0.01 and "Poisson" noise, where each pixel value is replaced by one drawn from a Poisson distribution with the mean equal to the original pixel value, to avoid excessive noise. To smooth the resulting images and reduce sharp edge effects, a 5x5 median filter was applied, creating a more realistic smooth gradient in the final images. This systematic approach ensures that the generated synthetic TEM images effectively encode the essential concepts required for a robust ML model training.

C. Pipeline and architectures.

The detailed architecture of the layers in both neural networks employed (i.e., Net1 & Net2) is given in Fig. 5.

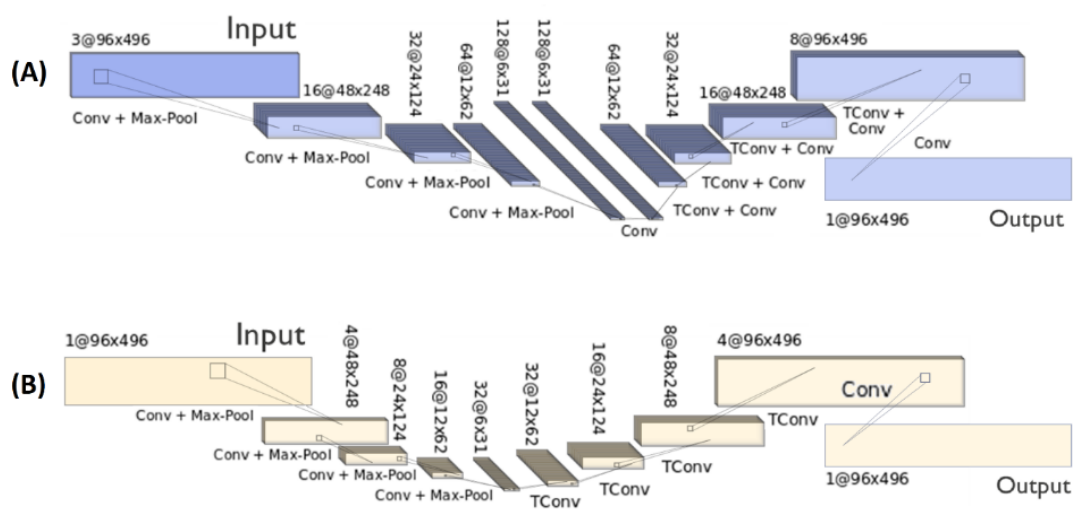

*Figure 5: Layers architecture for Net1 and Net2. (A) The detailed layered architecture for Net1. (B) The detailed layered architecture for Net2. Conv stands for convolution process. TConv stands for transverse convolution process. Max-Pool stand for maxpooling process to reduce dimensionality. The dimension for each layer input is given above each data stage in the format (No. of channels)@(image width in pixels)×(image length in pixels).*

Both networks are composed of an encoder part, where features extraction is done using four successive stages of convolution followed by max-pooling, and a decoder part where features are used to generate the output image using four successive stages of transverse convolution. However, for Net1 an extra convolution stage is used as a bridge between encoder and decoder to allow for an extra step of interconnections between extracted features before they proceed into the decoder part. An additional convolution step is incorporated into each transverse convolution stage in the decoder part of Net1, in contrast to Net2. These extra layers function as depth increasers in Net1, thereby augmenting the number of learnable parameters. This enhancement is designed to ensure the high quality of the initial guess produced by Net1, which is critical for the overall accuracy of our system in extracting the microstructure.

### D. Training

For the training of the first neural network, edge detection algorithms were applied to the generated 3,000 synthetic TEM images to produce the corresponding RGB images. These RGB images served as input for training Net1. The binary images, generated by our Monte Carlo module, were used as targets. After training Net1, it was used to produce 3,000 binary initial guess images from the RGB inputs, which were then utilized as inputs for training the second neural network Net2, again using the binary images as targets. For both training processes, the images were randomly shuffled and divided into three sets: 85% for training, 7.5% for validation, and 7.5% for testing. Training was conducted using MATLAB's deep learning toolbox with the Adaptive Moment Estimation (ADAM) optimizer [38]. Each convolutional layer employed a Rectified Linear Unit (ReLU) activation function, and batch normalization layers were inserted between each convolutional layer and its subsequent ReLU layer to accelerate the training process by allowing for higher learning rates and improving the generalization capabilities of our model [39]. An initial learning rate of 0.1 was used, with a learning rate reduction factor of 0.5 applied every 12 epochs. The batch size was set to 102 examples, resulting in 25 iterations of error minimization per epoch. Training continued until a prolonged period of stabilization was observed, accompanied by a slight increase in the validation set error, indicating entry into the overfitting regime of error minimization. The final trained network was selected as the one which achieved the minimum validation error to avoid overfitting. [40].

## Conclusion

This study provides an advancement in the data preparation stage of creating a ML model, where a novel concept-oriented synthetic data method is proposed which helps in tackling the challenge of data scarcity and provides a way to control the ML model behaviour through data creation from a set of concepts. This advancement in data preparation complements the recent advances in model architectures such as transformers with multi-head attention layers [41] and in the training processes such as physics informed machine learning [42]. The presented method was applied to the problem of extracting microstructure from different microscopy images of polycrystalline metal interconnects, where the framework integrating edge detection algorithms with generative ML diffusion models was demonstrated to effectively identify grain boundaries, achieving an average accuracy of 97.23%.

# Acknowledgements

Acknowledgements

The authors extend their sincere gratitude to Marleen van der Veen (Principal Member of Technical Staff at imec, Belgium) for generously providing the TEM images for Ruthenium (Ru) interconnects. Her support and contributions are greatly appreciated. The authors would also express their appreciation for the help provided by Olalla Varela Pedreira (Quality and Reliability Testing at imec, Belgium) and Patrick Carolan (Senior Process engineer imec, Belgium) during the preparation of the TEM images for Copper (Cu) interconnects. This work has been enabled in part by the NanoIC pilot line. The acquisition and operation are jointly funded by the Chips Joint Undertaking, through the European Union's Digital Europe (101183266) and Horizon Europe programs (101183277), as well as by the participating states Belgium (Flanders), France, Germany, Finland, Ireland and Romania. For more information, visit nanoic-project.eu.